# Heterogeneous network and graph attention auto-encoder for LncRNA-disease association prediction

Jin-Xing Liu, *Member, IEEE*, Wen-Yu Xi, Ling-Yun Dai, Chun-Hou Zheng, *Member, IEEE*, Ying-Lian Gao

***Abstract*—The emerging research shows that lncRNAs are associated with a series of complex human diseases. However, most of the existing methods have limitations in identifying nonlinear lncRNA-disease associations (LDAs), and it remains a huge challenge to predict new LDAs. Therefore, the accurate identification of LDAs is very important for the warning and treatment of diseases. In this work, multiple sources of biomedical data are fully utilized to construct characteristics of lncRNAs and diseases, and linear and nonlinear characteristics are effectively integrated. Furthermore, a novel deep learning model based on graph attention automatic encoder is proposed, called HGATELDA. To begin with, the linear characteristics of lncRNAs and diseases are created by the miRNA-lncRNA interaction matrix and miRNA-disease interaction matrix. Following this, the nonlinear features of diseases and lncRNAs are extracted using a graph attention auto-encoder, which largely retains the critical information and effectively aggregates the neighborhood information of nodes. In the end, LDAs can be predicted by fusing the linear and nonlinear characteristics of diseases and lncRNA. The HGATELDA model achieves an impressive AUC value of 0.9692 when evaluated using a 5-fold cross-validation indicating its superior performance in comparison to several recent prediction models. Meanwhile, the effectiveness of HGATELDA in identifying novel LDAs is further demonstrated by case studies. the HGATELDA model appears to be a viable computational model for predicting LDAs.***

## I. INTRODUCTION

LONG non-coding RNAs (lncRNAs) are a class of RNA molecules that are longer than 200 nucleotides and do not encode proteins [1-4]. Despite being previously classified as "junk DNA," recent research has demonstrated that long non-coding RNAs (lncRNAs) are vital in numerous biological processes, including gene expression regulation, chromatin modification, and epigenetic regulation. Additionally, dysregulated expression of lncRNAs has been linked to the onset and progression of diverse diseases, such as cancer, cardiovascular disease, and neurological disorders [5-7]. Identifying the associations between lncRNAs and diseases holds significant implications for comprehending disease mechanisms, devising novel diagnostic tools, and identifying potential therapeutic targets. However, further experimental validation is required to confirm these associations and establish their clinical relevance [8, 9]. Mining the potential LDAs is of far-reaching significance to the prevention and treatment of diseases, and to help medical staff understand the pathological mechanism of various complex diseases [10]. However, the wet experimental verification of LDAs is laborious and limited to small-scale. Therefore, efficient and economical computational models provide the conditions to predict potential LDAs on a large scale [11, 12].

In recent years, the prediction of lncRNA-disease associations has received significant attention from researchers both domestically and internationally [13, 14]. Despite significant progress in the field, challenges still remain, such as the lack of experimental validation and the need for more accurate computational methods. Nevertheless, the domain of predicting associations between long non-coding RNAs (lncRNAs) and diseases is rapidly expanding, with the potential to make significant contributions to our understanding of disease mechanisms and the development of innovative therapeutics [15, 16]. In conclusion, the development of a feasible predictive model is crucial for validating potential LDAs.

To date, numerous computational approaches have been proposed for LDA prediction [17-19], which can be broadly classified into three main groups: machine learning-based prediction, network-based prediction and deep learning-based prediction. The first category is the machine learning-based prediction. These methods utilize the similarity between lncRNA and disease profiles to predict potential lncRNA-disease

---

This work was supported in part by the grants provided by the National Natural Science Foundation of China, No. 62172254. *(Corresponding author: Ying-Lian Gao).*

Jin-Xing Liu is with the School of Health and Life Sciences, University of Health and Rehabilitation Sciences, Qingdao, Shandong, 276800 China, (e-mail: sdcavell@126.com), and with School of Computer Science, Qufu Normal University, Rizhao, Shandong.

Wen-Yu Xi is with the School of Computer Science, Qufu Normal University, Rizhao, 276826, China (e-mail: xiwenyu1007@126.com).

Ling-Yun Dai is with the School of Computer Science, Qufu Normal University, Rizhao, Shandong, 276826 China (e-mail: dailingyun_1@163.com).

Chun-Hou Zheng is with the School of Computer Science, Qufu Normal University, Rizhao 276826, China (e-mail: zhengch99@126.com).

Ying-Lian Gao is with the Library of Qufu Normal University, Qufu Normal University, Rizhao, Shandong, 276826 China (e-mail: yinliangao@126.com).

associations. For instance, Chen *et al*. proposed LRLSLDA, which employs Laplacian regularized least squares to predict unknown LDAs and achieves improved accuracy [20]. Xie *et al*. developed a model for predicting LDA based on a regularized least squares approach with normal Laplacian regularization, called SKF-LDA [21]. SKF-LDA adopted a more appropriate fusion method that incorporated more biological knowledge, resulting in more accurate predictions. Yu *et al*. proposed CFNBC model [22], which utilized Naïve Bayes classifier to predict potential LDAs. Lu *et al*. proposed SIMCLDA model [23], which utilizes principal component analysis and inductive matrix completion to predict potential LDAs. However, these methods do not leverage all the diverse data relevant to lncRNAs, which could affect the model's performance to a certain extent.

The second group is the network-based prediction. Sun *et al*. developed the random walk model (RWRlncD) on a known network to predict potential associations [24]. Zhang *et al*. proposed a model based on random walk, called BRWMC [25]. The model uses similarity network fusion and random walk methods to predict potential LDAs. Later, Hu *et al*. developed a model to predict LDA, called BiWalkLDA [26]. In this model, the scores of disease side and lncRNA side are obtained by double random walk algorithm, and the average score is used as the result. Gu *et al*. developed a model based on global network random walk (GrwLDA)[27], which has the greatest advantage of not requiring negative samples and enhancing the predictive performance of the model for LDAs. However, these methods are dependent on the known correlation information.

The third group is the deep learning-based prediction. With the development of research, the deep learning methods have been successfully extended to numerous aspects of life, such as natural language processing, computer vision, automatic driving [28], etc. Moreover, researchers have applied deep learning to biological network [29-31], and made it a reliable technology for LDA prediction. For example, Fan *et al*. proposed a method (GCRFLDA) based on the completion of the convolution matrix model [32], this model introduces attention mechanism and conditional random field to predict potential LDAs. Zhao *et al*. developed a novel prediction model called HGATLDA [33], which combines meta-path and graph attention model to predict LDAs. Ma *et al*. combined deep network fusion and graph embedding technology to establish a deep multi-network embedding model to predict potential LDAs [34]. Meanwhile, there is room for improvement in these models. In each model, only single-category features are utilized to forecast potential LDAs.

In this study, a new model is proposed to predict LDAs, called HGATELDA. In our model, we utilized multiple sources from biological data, and combined linear and nonlinear features to construct LncRNA and disease feature representation. Finally, a deep neural network is employed to obtain the prediction results. The flowchart of our model is presented in Fig. 1. Meanwhile, our model's advantages are manifested in the following aspects:

- To obtain more association data information, we make full use of biological characteristics from a variety of sources to construct features of diseases and lncRNAs, including diseases, miRNAs, lncRNAs.
- The linear characteristics of lncRNAs and diseases are established by the similarity matrix and miRNA correlation network, which largely retains the initial information.
- To extract the nonlinear characteristics of lncRNAs and diseases, the graph attention auto-encoder is utilized, which largely retains the critical information and effectively aggregates the neighborhood information of nodes.
- To obtain more informative representations, we combine multiple categories of features. By fusing the linear and nonlinear features of diseases and lncRNA, the potential LDAs can be predicted.
- Our comprehensive experimental results and case studies indicated that the performance of HGATELDA method outperforms other advanced methods. This further validates the contribution and efficacy of our model in predicting disease-related lncRNAs.

Other parts of this article are described below. After introducing the work related to lncRNAs and diseases (section 1), we give the experimental data and methods in sections 2 and 3. The experimental results are given in section 4. Finally, we concluded in section 5.

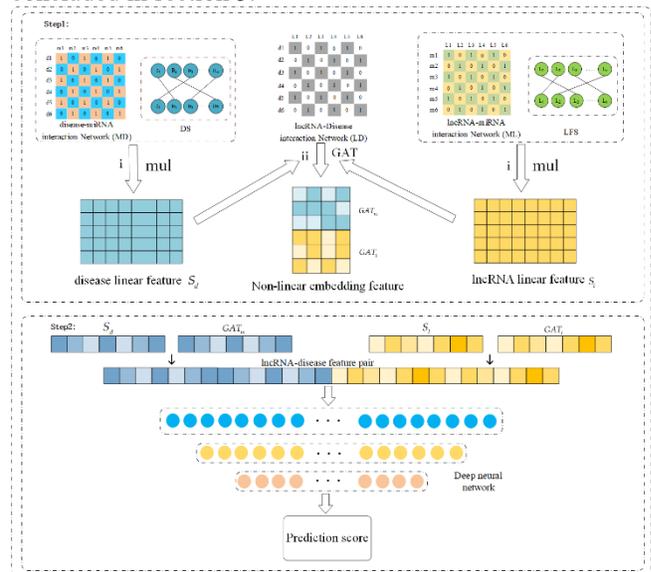

Fig. 1. The overall framework of HGATELDA. Step 1: ⅰ. the disease linear feature is obtained by linear multiplication of the disease-miRNA interaction matrix and disease semantic similarity matrix; The lncRNA linear feature is obtained by linear multiplication of the lncRNA-miRNA interaction matrix and the lncRNA functional similarity matrix. ⅱ. The nonlinear characteristics of disease and lncRNA are obtained by using a graph attention network. Step 2: The linear and nonlinear features of disease and lncRNA are fused to get the lncRNA-disease feature representation, which is input to the deep neural network for prediction.

## II. MATERIALS

### A. Datasets

In recent years, many public databases have been used to predict the LDAs. In this paper, a commonly used dataset, was obtained from Fu *et al.*'s study, including 240 lncRNAs, 495 miRNAs, and 412 diseases. Specifically, this information includes 2697 LDAs obtained from the Lnc2Cancer [35], and LncRNADisease [36], 13562 MDAs come from the HMDD (v2.0) database [37], and 1002 MLAs are obtained from the StarBase database [38]. The dataset details are showed in Table 1.

TABLE I
THE DATASET DETAILS USED IN THE EXPERIMENT

| Datasets | Data sources | Amounts |
|---|---|---|
| LDA | LncRNADisease databases; lnc2Cancer databases | 2697 |
| MDA | HMDD (v2.0) | 13562 |
| MLA | StarBase | 1002 |

### B. Disease Semantic Similarity

In this section, through directed acyclic graph ($DAG$) of disease to depict their relationships and compute their semantic similarity. Consequently, for a disease D, its semantic value can be expressed as:

$$DV(D) = \sum_{d \in T(D)} D_D(d), \quad (1)$$

f disease $D$ and contains all ancestor nodes. Consequently, the semantic contribution value of a disease $m$ is calculated by:

$$\begin{cases} D_d(m)=1 & if\ m=d, \\ D_d(m) = \max\{\Delta * D_d(\tilde{m}) | \tilde{m} \in children\ of\ m\} & if\ m \neq d, \end{cases} \quad (2)$$

where $\Delta$ is the semantic contribution factor. In general, $\Delta = 0.5$ [39]. After that, the similarity between the two diseases $d_i$ and $d_j$ is computed as follows:

$$DS(d_i, d_j) = \frac{\sum_{m \in T(A) \cap T(B)} \left( D_{d_i}(m) + D_{d_j}(m) \right)}{DV(d_i) + DV(d_j)}. \quad (3)$$

### C. LncRNA Functional Similarity

Based on the diseases with similar phenotypes are usually associated with functionally similar lncRNAs. Here, we employed the model proposed by Chen et al. to assess the functional similarity of lncRNA pairs by measuring the semantic similarity between two disease-related lncRNA groups [40]. Therefore, LFS can be defined as:

$$LFS = \frac{\sum_{d \in D(l_j)} S(d, D(l_i)) + \sum_{d \in D(l_i)} S(d, D(l_j))}{m+n}, (4)$$

$$S(d, D(l_i)) = \max_{d1 \in D(l_i)} (S(d, d_1)), \quad (5)$$

where m and n represent diseases numbers of lncRNA $l_i$ and $l_j$, and $D(l)$ represents the diseases related to lncRNA $l$. Eq.5 computes the similarity between a disease element $d$ in the set $D(l_i)$ and the entire set $D(l_i)$.

## III. METHODOLOGY

### A. The Construction of Linear Features

To obtain more similarity information, we introduce miRNA data into the model. In this case, linear multiplication is utilized to extract the linear characteristics of lncRNAs and diseases, thereby obtaining additional node similarity information. The linear feature of lnRNA is obtained by multiplying $LFS$ and MLA profiles $ML$:

$$FL = LFS \times ML, \quad (6)$$

Meanwhile, the linear feature of disease is obtained by multiplying disease semantic similarity $DS$ and miRNA-disease interaction profiles $MD$:

$$FD = DS \times MD. \quad (7)$$

An l-dimensional vector can then be constructed for each disease or lncRNA. Finally, the features of all diseases and lncRNA are represented by F:

$$F = \begin{bmatrix} FL \\ FD \end{bmatrix} = \begin{bmatrix} f_1 \\ f_2 \\ ... \\ f_{m+n} \end{bmatrix} \in R^{(p+q)l}, \quad (8)$$

where $p+q$ denotes the total number of nodes, with $f \in R^l$ denotes the linear characteristics of each node.

### B. The construction of non-linear features

As an unsupervised learning model, Graph Attention Autoencoder (GATE) can reconstruct the node attributes of data through the encoder and decoder. Fig. 2 shows the specific architecture of GATE. Meanwhile, GATE calculates the importance of nodes' neighbors through attention mechanism and updates the characteristics of nodes.

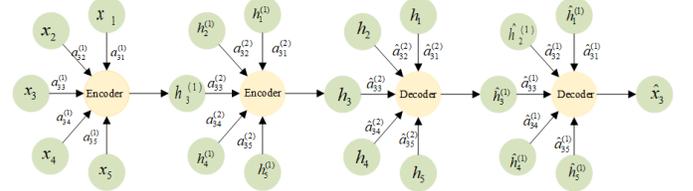

Fig.2. The process of using GATE to reconstruct the features of node.

At first, the lncRNA-disease graph by the association matrix LD, and it is represented by $G=(N,E)$. $N = \{n_1, n_2, ..., n_m + n_n\}$ represents the vertices, $E$ is defined as the edge between nodes, and the original characteristics of nodes in graph G are denoted by $F$. Following that, the attention mechanism is applied to determine the contribution value of each node and its neighbors. To be more specific, the attention coefficient $z_{ij}$ between node $e_i$ and its neighbor $e_j$ can be computed as:

$$z_{ij}(e_i, e_j) = leakyReLu(\alpha^T \left[ Wf_i \| Wf_j \right]), \quad (9)$$

where $W \in R^{l' \times l}$ serves as a transformation matrix that projects the initial node characteristics into a space with l' dimensions, and *leakyReLu* refers to a nonlinear activation function. $\alpha \in R^{2l'}$ represents the attention parameter, which assigns real numbers to features.

To eliminate dimensionality between different attention coefficients, the attention coefficients $z_{ij}$ are further normalized as follows:

$$\sigma_{ij} = \text{soft}\max(z_{ij}) = \frac{\exp(z_{ij})}{\sum_{t \in N_i}\exp(z_{ij})}, \quad (10)$$

where $E_i$ represents the set of neighboring nodes of node $e_i$. $\sigma_{ij}$ indicates the importance of node $e_i$ in relation to node $e_j$ based on the normalized attention coefficient.

To update the representation of the given node $e_i$, the concentration coefficients calculate the importance of information from its neighboring nodes, and then aggregate this information. This aggregation process typically involves a weighted sum of the neighboring node features, with the weights determined by the concentration coefficients. By using these coefficients to aggregate information from the neighbors, the representation of the given node can be updated to incorporate information:

$$f'_i = \xi(\sum_{t \in N_i}\sigma_{it}Wf_t), \quad (11)$$

where $\xi$ is the *leakyReLu* activation function.

Multi-head attention enhances the stability of the self-attention learning process and strengthens the model's ability to extract information by mitigating biases. In addition, to enhance information capture capability, multi-head attention is used to pick up information from different representation spaces, thus enhancing its learning capacity. The integration of the K-independent attention mechanism is specifically carried out in the following manner:

$$f'_i = \xi(\frac{1}{K}\sum_{k=1}^{K}\sum_{t \in E_i}\sigma_{it}^k \cdot W^k f_t). \quad (12)$$

The output of a Graph Attention Layer is:

$$F' = \begin{bmatrix} f'_1 \\ f'_2 \\ \ldots \\ f'_{m+n} \end{bmatrix} \in R^{(m+n)l'}. \quad (13)$$

A graph consisting of lncRNA-disease associations is inputted into a graph attention autoencoder along with the linear characteristics F of each node. The propagation of features and the fusion of attention allows for obtaining non-linear representations of the nodes.

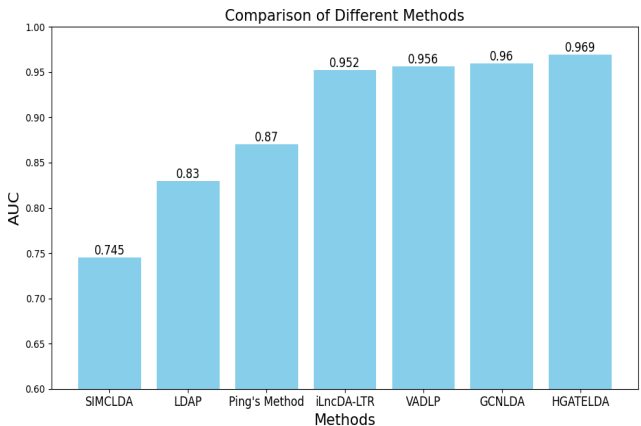

Fig.3. Comparison of AUC values with different methods.

### C. Fusion lncRNA-disease Features

Existing methods typically utilize either linear or nonlinear features for lncRNA-disease prediction, but such single-category features are insufficient to capture the complex relationships between them. To solve the above problem, the linear and nonlinear features are integrated into our model for prediction. Particularly, new lncRNA feature vectors and disease node feature vectors are obtained by connecting linear and nonlinear features:

$$Lnc = [FL, FL'], \quad (14)$$

$$Dis = [FD, FD']. \quad (15)$$

Subsequently $F_{ld}$ represents the characteristics of the lncRNA-disease pair $(i,j)$ as follows:

$$F_{ld} = [f_{li}, f'_{li}, f_{dj}, f'_{dj}] \in R^{2 \times (l+l')}. \quad (16)$$

## IV. RESULTS AND DISCUSSION

### A. Evaluation Metrics

To assess the performance of HGATELDA, we employ the receiver operating characteristic (ROC) curve as the evaluation metric and calculate the area under the curve (AUC) of the ROC. Additionally, three evaluation metrics, namely Accuracy (ACC), F1-Score (F1), Precision (Pre) and Matthews correlation coefficient (Mcc) are also computed, as follows:

$$Acc = \frac{TP+TN}{TP+FN+TN+FP}, \quad (17)$$

$$\text{Recall} = \frac{TP}{TP+FN}, \quad (18)$$

$$F1 = \frac{2TP}{2TP+FP+FN}, \quad (19)$$

$$Mcc = \frac{TP \times TN - FP \times FN}{\sqrt{(TP+FP)(TP+TN)(TN+FP)(TN+FN)}}, \quad (20)$$

$$\text{Pre} = \frac{TP}{TP+FP}. \quad (21)$$

### B. Comparative Experiment

To demonstrate the predictive capabilities of HGATELDA, it underwent a 5-fold cross-validation process, where it was compared against various other methods: SIMCLDA [23], LDAP [41], Ping's method [42], iLncDA-LTR [16], VADLP [43] and GCNLDA [44]. As shown in Fig. 3, our model achieved the best AUC value of 0.969, which was 0.4% higher than the GCNLDA, 1.3% better than VADLP, 1.7% better than iLncDA-LTR, 9.9% higher than Ping's method, 13.9% and 22.4% higher than LDAP and SIMCLDA.

The factors that have achieved good experimental results are described as follows: To begin with, the linear characteristics of lncRNAs and diseases are established by the MLI matrix and MDA matrix. Following this, the graph attention auto-encoder extracts the nonlinear features of diseases and long noncoding RNAs, largely retaining the critical information. In the end, by combining both linear and nonlinear characteristics of diseases and lncRNAs. The results of the 10-fold cross-validation of HGATELDA are presented in Table 2 and Fig. 4. Subsequently,

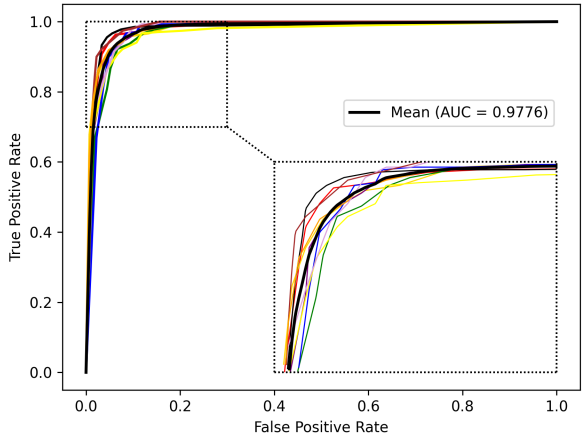

Fig.4. Ten-fold CV curve of HGATELDA.

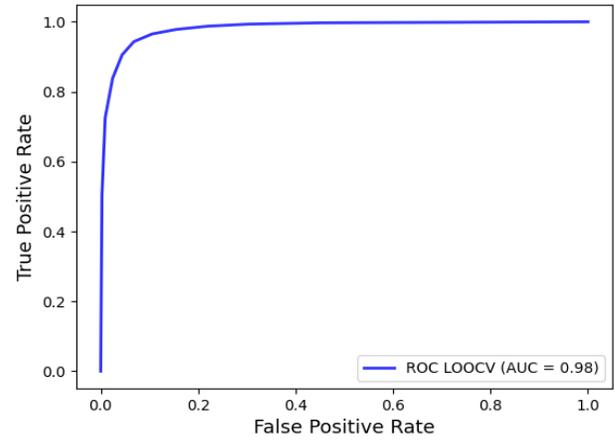

Fig.5. LOOCV curve of HGATELDA.

TABLE Ⅱ
TEN-FOLD CV RESULTS PERFORMED BY HGATELDA

| Fold | Acc | Sen | Spec | Pre | Mcc |
|---|---|---|---|---|---|
| 0 | 0.946 | 0.963 | 0.930 | 0.932 | 0.893 |
| 1 | 0.939 | 0.967 | 0.911 | 0.916 | 0.879 |
| 2 | 0.952 | 0.956 | 0.948 | 0.949 | 0.904 |
| 3 | 0.943 | 0.941 | 0.944 | 0.944 | 0.885 |
| 4 | 0.930 | 0.952 | 0.907 | 0.911 | 0.860 |
| 5 | 0.920 | 0.937 | 0.904 | 0.907 | 0.841 |
| 6 | 0.939 | 0.948 | 0.930 | 0.931 | 0.878 |
| 7 | 0.931 | 0.937 | 0.926 | 0.927 | 0.863 |
| 8 | 0.948 | 0.967 | 0.930 | 0.932 | 0.897 |
| 9 | 0.941 | 0.944 | 0.937 | 0.938 | 0.881 |
| Average | 0.939 | 0.951 | 0.927 | 0.928 | 0.878 |

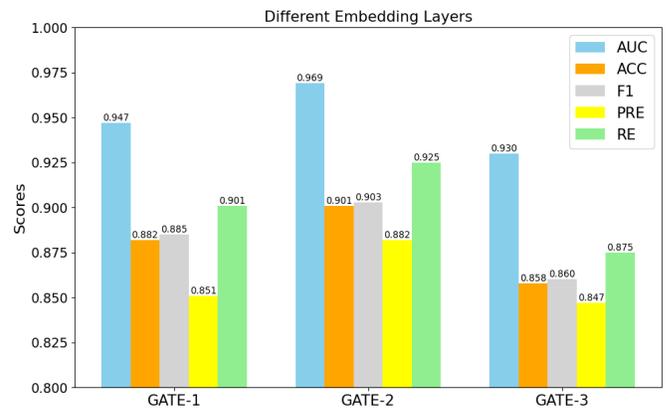

Fig.6. AUC under different numbers of talking heads k.

Fig.5 depicts the results of leave-one-out cross-validations (LOOCV).

Here, the performance of the model is proved by comparing different feature combinations. At first, the linear feature is obtained by basic linear multiplication (combination 1). Secondly, the nonlinear characteristics of prediction are obtained by GAT (combination 2). Finally, the combination of linear features and nonlinear features is used to obtain the prediction feature (combination 3).

The specific results are shown in Table 3, which shows that combination 3 has achieved the best experimental results. Based on the results, we can conclude that combining linear and nonlinear features can yield a greater amount of information and improve the model's performance.

At the same time, we study the layers of decoder and encoder of the GATE model, abbreviated as GATE-1, GATE-2 and GATE-3. All the evaluation results are recorded in Fig. 6. The experimental results show that the best experimental results are obtained when the number of layers of decoder and encoder is set to 2. In our study, we investigated the impact of the number of attention heads in multi-head attention on the model's performance. Fig. 7 clearly illustrates the trend of how different numbers of heads affect the model's performance. This research helps us gain a better understanding of how parameter selection within the model affects performance, offering valuable insights for optimizing the model.

The performance of HGATELDA will vary based on the parameter values. There are many hyperparameters in the DNN classifier that needs to be tuned as follows:

(1) Number of layers and hidden neurons. The model employs a three-layer neural network, consisting of 128, 64, and 32 neurons, respectively.

(2) Optimizer and learning rate. The classifier utilizes the Adam optimizer, and the learning rate is set to $10^{-3}$.

(3) Dropout. To prevent overfitting of the model, we carried out sensitivity analysis of dropout $\alpha$. As shown in Fig. 8, the best experimental results are obtained when $\alpha = 0.2$.

(4) Weights and biases are initialized to 0 as the initial values. The hyperparameters for GATE are configured as follows:

(1) Number of layers: Both the encoder and decoder are set

TABLE Ⅲ
COMPARATIVE EXPERIMENT OF FEATURE FUSION

| Feature fusion | AUC | ACC | F1 | PRE | RE |
|---|---|---|---|---|---|
| combination 1 | 0.914 | 0.819 | 0.821 | 0.829 | 0.813 |
| combination 2 | 0.933 | 0.860 | 0.865 | 0.836 | 0.895 |
| combination 3 | 0.969 | 0.901 | 0.903 | 0.882 | 0.925 |

to 2 layers.

(2) Number of neurons per layer: The decoder layer aligns with the corresponding encoder layer, with 128 and 64 neurons for the two encoder layers, respectively.

(3) Learning rate, Lambda, and Dropout: We set the learning rate to 0.001, enabling fast convergence of the model.

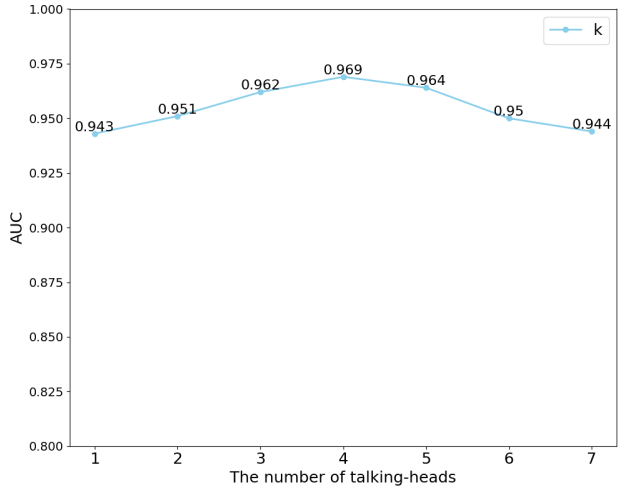

Fig.7. AUC under different numbers of talking heads k.

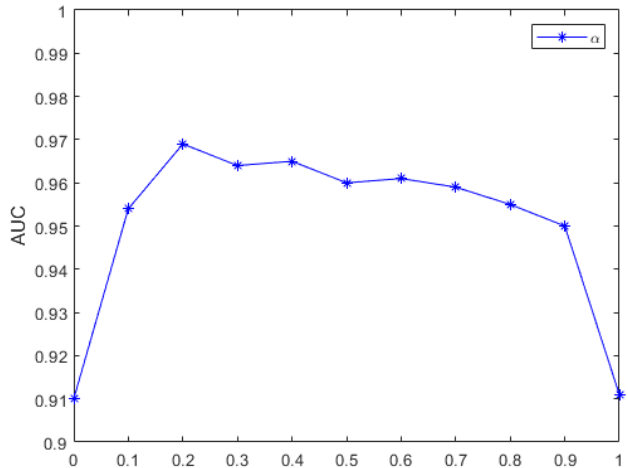

Fig.8. AUC under different dropout rate $\alpha$.

### C. Case Study

This section provides case studies of three types of human cancers to further illustrate the predictive capabilities of GATELDA. The cancers studied include breast, pancreatic, and colorectal cancer. The process involves ranking candidate lncRNAs for each disease based on their final predicted score. The top lncRNA for each cancer is then selected for analysis in two databases: Lnc2cancer (I) and LncRNADisease (II). Here, the candidate genes labeled as "literature" are all confirmed to be related to pancreatic cancer in the literature.

The first disease of choice is breast cancer. Breast cancer is a common malignant tumor in gynecology, which seriously endangers the physical and mental health of women.

TABLE IV
PREDICTED LNCRNAS FOR BREAST CANCER

| Rank | lncRNA | Evidence | PMID |
|---|---|---|---|
| 1 | TUG1 | I; II | 35232340 |
| 2 | HULC | I; II | 37910386 |
| 3 | MIR17HG | literature | 36943627 |
| 4 | BANCR | I; II | 29565494 |
| 5 | PCAT1 | I; II | 35014684 |
| 6 | PRNCR1 | II | 33608112 |
| 7 | HNF1A-AS1 | literature | 33603481 |
| 8 | GHET1 | I; II | 30787968 |
| 9 | TUSC7 | literature | 34305410 |
| 10 | LINC01133 | I; II | 31557401 |
| 11 | GHET1 | I; II | 30787968 |
| 12 | SOX2-OT | I; II | 34997317 |
| 13 | FAL1 | I; II | 29987852 |
| 14 | PVT1 | I; II | 37531833 |
| 15 | CDKN2B-AS1 | I; II | 35965791 |

GATELDA was used to predict the 15 potential lncRNA with the highest correlation score with breast cancer to analyze whether they were associated. The results are listed in Table 4. Out of the initial 15 new lncRNA predictions, 12 have been linked to breast cancer in the database, indicating a high success rate for these potential lncRNA. It's important to note that the validity of these lncRNA has been confirmed by two databases, further emphasizing their potential significance in prostate cancer research. For example, Recent research has found that the TUG1 gene is abnormally expressed in breast cancer cells [45]. This study revealed that TUG1 may play a significant role in the development of breast cancer, as its abnormal expression may lead to cell proliferation and metastasis [46]. This discovery provides new directions for further research and holds promise for the development of more effective methods to treat breast cancer. Although MIR17HG, HNF1A-AS1 and TUSC7 are not recorded in the database, they are all supported by the literature [47-49]. The treatment of breast cancer is significantly impacted by the above three types of lncRNAs, as indicated by recent research findings.

The second disease of choice is pancreatic cancer. Due to its hidden and atypical nature, pancreatic cancer poses significant challenges to both diagnosis and treatment, and its incidence and mortality rates are among the highest in malignant tumors. In this context, we have selected 15 potential lncRNAs with the highest correlation scores with pancreatic cancer to investigate their potential association with pancreatic cancer, and the results are listed in Table 5. Out of the initial 15 newly predicted lncRNAs, 13 have been found to be associated with pancreatic cancer in the database. It is worth noting that 11 out of these potential lncRNAs have been validated by two separate databases. For instance, a study reported a significant increase in the expression level of BANCR in pancreatic cancer tissues, where high expression was correlated with advanced clinical stage and poor patient prognosis [50]. A separate study discovered that BANCR can regulate the expression of miR-195 to promote the proliferation and invasion of pancreatic cancer cells [51]. These studies suggest that BANCR may be a potential therapeutic target and biomarker for pancreatic cancer. Among them, the study found that HNF1A-AS1, ADPGK-AS1 and PCA3 are related to the clinicopathological features of pancreatic cancer, which indicate that three kinds of lncRNAs are potential therapeutic targets for pancreatic cancer [52-54].

TABLE VI
PREDICTED LNCRNAS FOR COLORECTAL CANCER

| Rank | lncRNA | Evidence | PMID |
|---|---|---|---|
| 1 | MIR17HG | I; II | 34145399 |
| 2 | PCAT1 | I; II | 31253700 |
| 3 | SPRY4-IT1 | I; II | 33029299 |
| 4 | HNF1A-AS1 | I; II | 36230970 |
| 5 | IGF2-AS | literature | 32853944 |
| 6 | CDKN2B-AS1 | literature | 34436551 |
| 7 | BCYRN1 | I | 31773686 |
| 8 | WT1-AS | literature | 30714675 |
| 9 | MIR100HG | I; II | 35279145 |
| 10 | PCA3 | literature | 25381039 |
| 11 | CRNDE | I; II | 37716979 |
| 12 | SOX2-OT | literature | 37524903 |
| 13 | LINC00675 | I; II | 29524886 |
| 14 | LINC01133 | I; II | 38146619 |
| 15 | HCP5 | I; II | 32606965 |

The last disease of choice is colorectal cancer. Colorectal cancer ranks third among the most prevalent malignant tumors globally and is the second leading cause of death. Early detection through timely diagnosis increases the likelihood of successful treatment through surgery, resulting in a higher cure rate. Similarly, the top 15 results are chosen. The results are listed in Table 6. In colorectal cancer samples, CRNDE has been found to be upregulated in colorectal cancer tissues and has been associated with tumor growth and progression [55]. However, further research is needed to fully understand the role of CRNDE in colorectal cancer. In addition, PCAT1 has been found to be upregulated in colorectal cancer tissues and has been associated with tumor growth and progression [56]. Although there are five lncRNAs that have not been confirmed, experimental results indicate that these five lncRNAs are also closely related to prostate cancer, and it has been confirmed in the literature that these five lncRNAs are also closely related to colorectal cancer [57, 58].

TABLE V
PREDICTED LNCRNAS FOR PANCREATIC CANCER

| Rank | lncRNA | Evidence | PMID |
|---|---|---|---|
| 1 | TUG1 | I; II | 33026604 |
| 2 | BANCR | I; II | 36284647 |
| 3 | HNF1A-AS1 | literature | 32110048 |
| 4 | ADPGK-AS1 | literature | 29667486 |
| 5 | XIST | I | 28295543 |
| 6 | NEAT1 | I; II | 36951525 |
| 7 | TUSC7 | I | 30714151 |
| 8 | CASC2 | literature | 28865121 |
| 9 | UCA1 | I; II | 36741256 |
| 10 | PCAT1 | I; II | 36093435 |
| 11 | DANCR | I; II | 31515968 |
| 12 | LINC-PINT | I; II | 30944652 |
| 13 | H19 | I; II | 32272875 |
| 14 | LINC00663 | I; II | 35647035 |
| 15 | GHET1 | I; II | 29228419 |

## V. CONCLUSION

The growing body of evidence indicates that the identification of potential LDAs holds immense importance in both comprehending disease pathogenesis and advancing clinical medicine. As biomedical data continues to grow in complexity and size, new methods are needed to effectively integrate and analyze this information. In our study, we propose a novel computational model that utilizes a heterogeneous network and graph attention auto-encoder to attain a deeper comprehension of candidate LDAs. Initially, we construct the linear characteristics of lncRNAs and diseases based on various biological premises related to lncRNAs, diseases, and miRNAs. Subsequently, the nonlinear features of diseases and lncRNAs are extracted using a graph attention auto-encoder. Finally, the linear and nonlinear characteristics of both diseases and lncRNAs are combined to make predictions about their association. The predictive performance of HGATELDA is evaluated using 5-fold CV and its results are compared with those of other experiments. The effectiveness of the proposed method in predicting potential LDAs, including many unknown associations, is demonstrated in case study. Although our method has achieved high performance, due to data limitations, the proposed model did not incorporate additional data sources related to lncRNAs and diseases. At the same time, we will explore more interpretable models in the future to delve into deeper associations between lncRNAs and diseases.

## REFERENCES

[1] Y. Zhang, Y. Tao, and Q. Liao, "Long noncoding RNA: a crosslink in biological regulatory network," *Briefings in Bioinformatics,* vol. 19, no. 5, pp. 930-945, Sep, 2018.

[2] T. Derrien, R. Johnson, G. Bussotti, A. Tanzer, and R. Guigó, "The GENCODE v7 catalog of human long noncoding RNAs: Analysis of their gene structure, evolution, and expression," *Genome Research,* vol. 22, no. 9, pp. 1775-1789, 2012.

[3] K. C. Wang, and H. Y. Chang, "Molecular mechanisms of long noncoding RNAs," *Molecular Cell,* vol. 43, no. 6, pp. 904-914, 2011.

[4] M. Guttman, and J. L. Rinn, "Modular regulatory principles of large non-coding RNAs," *Nature,* vol. 482, no. 7385, pp. 339-46, 2012.

[5] X. Zeng, L. Liu, L. Lü, and Q. Zou, "Prediction of potential disease-associated microRNAs using

structural perturbation method," *Bioinformatics,* vol. 34, no. 14, pp. 2425-2432, 2018.

[6] M. Niu, Q. Zou, and C. Wang, "GMNN2CD: identification of circRNA–disease associations based on variational inference and graph Markov neural networks," *Bioinformatics,* vol. 38, no. 8, pp. 2246-2253, 2022.

[7] W. Tang, S. Wan, Z. Yang, A. E. Teschendorff, and Q. Zou, "Tumor origin detection with tissue-specific miRNA and DNA methylation markers," *Bioinformatics,* vol. 34, no. 3, pp. 398-406, 2018.

[8] R. J. Taft, K. C. Pang, T. R. Mercer, M. Dinger, and J. S. Mattick, "Non‐coding RNAs: regulators of disease," *The Journal of Pathology: A Journal of the Pathological Society of Great Britain and Ireland,* vol. 220, no. 2, pp. 126-139, 2010.

[9] M. Esteller, "Non-coding RNAS in human diseases," *Nature Reviews Genetics,* vol. 12, no. 12, pp. 861-874, 2011.

[10] H. Feng, D. Jin, J. Li, Y. Li, Q. Zou, and T. Liu, "Matrix reconstruction with reliable neighbors for predicting potential MiRNA–disease associations," *Briefings in Bioinformatics,* vol. 24, no. 1, pp. bbac571, 2023.

[11] Z. Cui, Y.-L. Gao, J.-X. Liu, J. Wang, J. Shang, and L.-Y. Dai, "The computational prediction of drug-disease interactions using the dual-network L 2, 1-CMF method," *BMC bioinformatics,* vol. 20, pp. 1-10, 2019.

[12] X. Chen, C. C. Yan, X. Zhang, and Z.-H. You, "Long non-coding RNAs and complex diseases: from experimental results to computational models," *Briefings in Bioinformatics,* vol. 18, no. 4, pp. 558-576, Jul, 2017.

[13] M.-M. Yin, J.-X. Liu, Y.-L. Gao, X.-Z. Kong, and C.-H. Zheng, "NCPLP: A Novel Approach for Predicting Microbe-Associated Diseases With Network Consistency Projection and Label Propagation," *IEEE Transactions on Cybernetics,* vol. 52, no. 6, pp. 5079-5087, Jun, 2022.

[14] F. Zhou, M.-M. Yin, C.-N. Jiao, J.-X. Zhao, C.-H. Zheng, and J.-X. Liu, "Predicting miRNA-disease associations through deep autoencoder with multiple kernel learning," *IEEE Transactions on Neural Networks and Learning Systems,* vol. 34, no. 9, pp. 5570 - 5579, 2021.

[15] Y. Li, J. Li, and N. Bian, "DNILMF-LDA: prediction of lncRNA-disease associations by dual-network integrated logistic matrix factorization and Bayesian optimization," *Genes,* vol. 10, no. 8, pp. 608, 2019.

[16] H. Wu, Q. Liang, W. Zhang, Q. Zou, A. E.-L. Hesham, and B. Liu, "iLncDA-LTR: Identification of lncRNA-disease associations by learning to rank," *Computers in Biology and Medicine,* vol. 146, pp. 105605, 2022.

[17] L. Wong, L. Wang, Z.-H. You, C.-A. Yuan, Y.-A. Huang, and M.-Y. Cao, "GKLOMLI: a link prediction model for inferring miRNA–lncRNA interactions by using Gaussian kernel-based method on network profile and linear optimization algorithm," *BMC bioinformatics,* vol. 24, no. 1, pp. 188, 2023.

[18] K. Zheng, X.-L. Zhang, L. Wang, Z.-H. You, B.-Y. Ji, X. Liang, and Z.-W. Li, "SPRDA: a link prediction approach based on the structural perturbation to infer disease-associated Piwi-interacting RNAs," *Briefings in Bioinformatics,* vol. 24, no. 1, pp. bbac498, 2023.

[19] L. Wang, L. Wong, Z.-H. You, and D.-S. Huang, "AMDECDA: Attention Mechanism Combined with Data Ensemble Strategy for Predicting CircRNA-Disease Association," *IEEE Transactions on Big Data*, pp. 1 - 11, 2023.

[20] X. Chen, and G.-Y. Yan, "Novel human lncRNA–disease association inference based on lncRNA expression profiles," *Bioinformatics,* vol. 29, no. 20, pp. 2617-2624, 2013.

[21] G. Xie, T. Meng, Y. Luo, and Z. Liu, "SKF-LDA: similarity kernel fusion for predicting lncRNA-disease association," *Molecular Therapy-Nucleic Acids,* vol. 18, pp. 45-55, 2019.

[22] J. Yu, Z. Xuan, X. Feng, Q. Zou, and L. Wang, "A novel collaborative filtering model for LncRNA-disease association prediction based on the Naïve Bayesian classifier," *BMC bioinformatics,* vol. 20, pp. 1-13, 2019.

[23] C. Lu, M. Yang, F. Luo, F.-X. Wu, M. Li, Y. Pan, Y. Li, and J. Wang, "Prediction of lncRNA–disease associations based on inductive matrix completion," *Bioinformatics,* vol. 34, no. 19, pp. 3357-3364, 2018.

[24] J. Sun, H. Shi, Z. Wang, C. Zhang, L. Liu, L. Wang, W. He, D. Hao, S. Liu, and M. Zhou, "Inferring novel lncRNA–disease associations based on a random walk model of a lncRNA functional similarity network," *Molecular BioSystems,* vol. 10, no. 8, pp. 2074-2081, 2014.

[25] G.-Z. Zhang, and Y.-L. Gao, "BRWMC: Predicting lncRNA-disease associations based on bi-random walk and matrix completion on disease and lncRNA networks," *Computational Biology and Chemistry,* vol. 103, pp. 107833, 2023.

[26] J. Hu, Y. Gao, J. Li, Y. Zheng, J. Wang, and X. Shang, "A novel algorithm based on bi-random walks to identify disease-related lncRNAs," *BMC bioinformatics,* vol. 20, pp. 1-11, 2019.

[27] C. Gu, B. Liao, X. Li, L. Cai, Z. Li, K. Li, and J. Yang, "Global network random walk for predicting potential human lncRNA-disease associations," *Scientific reports,* vol. 7, no. 1, pp. 12442, 2017.

[28] T. Yang, L. Hu, C. Shi, H. Ji, X. Li, and L. Nie, "HGAT: Heterogeneous graph attention networks for semi-supervised short text classification," *ACM Transactions on Information Systems (TOIS),* vol. 39, no. 3, pp. 1-29, 2021.

[29] Z. Zhang, J. Xu, Y. Wu, N. Liu, Y. Wang, and Y. Liang, "CapsNet-LDA: predicting lncRNA-disease associations using attention mechanism and capsule network based on multi-view data," *Briefings in Bioinformatics,* vol. 24, no. 1, pp. bbac531, 2023.

[30] Q. Liang, W. Zhang, H. Wu, and B. Liu, "LncRNA-disease association identification using graph auto-encoder and learning to rank," *Briefings in Bioinformatics,* vol. 24, no. 1, pp. bbac539, 2023.

[31] Q.-W. Wu, J.-F. Xia, J.-C. Ni, and C.-H. Zheng, "GAERF: predicting lncRNA-disease associations by graph auto-encoder and random forest," *Briefings in bioinformatics,* vol. 22, no. 5, pp. bbaa391, 2021.

[32] Y. Fan, M. Chen, and X. Pan, "GCRFLDA: scoring lncRNA-disease associations using graph convolution matrix completion with conditional random field," *Briefings in Bioinformatics,* vol. 23, no. 1, pp. bbab361, 2022.

[33] X. Zhao, X. Zhao, and M. Yin, "Heterogeneous graph attention network based on meta-paths for lncrna–disease association prediction," *Briefings in Bioinformatics,* vol. 23, no. 1, pp. bbab407, 2022.

[34] Y. Ma, "DeepMNE: deep multi-network embedding for lncRNA-disease association prediction," *IEEE Journal of Biomedical and Health Informatics,* vol. 26, no. 7, pp. 3539-3549, 2022.

[35] Y. Gao, P. Wang, Y. Wang, X. Ma, H. Zhi, D. Zhou, X. Li, Y. Fang, W. Shen, Y. Xu, S. Shang, L. Wang, L. Wang, S. Ning, and X. Li, "Lnc2Cancer v2.0: updated database of experimentally supported long non-coding RNAs in human cancers," *Nucleic Acids Research,* vol. 47, no. D1, pp. D1028-D1033, Jan 8, 2019.

[36] Z. Bao, Z. Yang, Z. Huang, Y. Zhou, Q. Cui, and D. Dong, "LncRNADisease 2.0: an updated database of long non-coding RNA-associated diseases," *Nucleic Acids Research,* vol. 47, no. D1, pp. D1034-D1037, Jan 8, 2019.

[37] Y. Li, C. Qiu, J. Tu, B. Geng, J. Yang, T. Jiang, and Q. Cui, "HMDD v2.0: a database for experimentally supported human microRNA and disease associations," *Nucleic Acids Research,* vol. 42, no. D1, pp. D1070-D1074, Jan, 2014.

[38] J.-H. Li, S. Liu, H. Zhou, L.-H. Qu, and J.-H. Yang, "starBase v2.0: decoding miRNA-ceRNA, miRNA-ncRNA and protein-RNA interaction networks from large-scale CLIP-Seq data," *Nucleic Acids Research,* vol. 42, no. D1, pp. D92-D97, Jan, 2014.

[39] J. X. Liu, Z. Cui, Y. L. Gao, and X. Z. Kong, "WGRCMF: A Weighted Graph Regularized Collaborative Matrix Factorization Method for Predicting Novel LncRNA-Disease Associations," *IEEE Journal of Biomedical and Health Informatics,* vol. PP, no. 99, pp. 1-1, 2020.

[40] X. Chen, C. Clarence Yan, C. Luo, W. Ji, Y. Zhang, and Q. Dai, "Constructing lncRNA functional similarity network based on lncRNA-disease associations and disease semantic similarity," *Scientific reports,* vol. 5, no. 1, pp. 1-12, 2015.

[41] W. Lan, M. Li, K. Zhao, J. Liu, F.-X. Wu, Y. Pan, and J. Wang, "LDAP: a web server for lncRNA-disease association prediction," *Bioinformatics,* vol. 33, no. 3, pp. 458-460, 2017.

[42] P. Ping, L. Wang, L. Kuang, S. Ye, M. F. B. Iqbal, and T. Pei, "A Novel Method for LncRNA-Disease Association Prediction Based on an lncRNA-Disease Association Network," *IEEE/ACM Transactions on Computational Biology and Bioinformatics,* vol. 16, no. 2, pp. 688-693, Mar-Apr, 2019.

[43] N. Sheng, H. Cui, T. Zhang, and P. Xuan, "Attentional multi-level representation encoding based on convolutional and variance autoencoders for lncRNA–disease association prediction," *Briefings in Bioinformatics,* vol. 22, no. 3, pp. bbaa067, 2021.

[44] P. Xuan, S. Pan, T. Zhang, Y. Liu, and H. Sun, "Graph convolutional network and convolutional neural network based method for predicting lncRNA-disease associations," *Cells,* vol. 8, no. 9, pp. 1012, 2019.

[45] S. Fan, Z. Yang, Z. Ke, K. Huang, N. Liu, X. Fang, and K. Wang, "Downregulation of the long non-coding RNA TUG1 is associated with cell proliferation, migration, and invasion in breast cancer," *Biomedicine & pharmacotherapy,* vol. 95, pp. 1636-1643, 2017.

[46] S. Mashhadizadeh, M. Tavangar, A. F. Javani, M. D. Rahimian, M. Azadeh, H. Tabatabaeian, and K. Ghaedi, "PGR and TUG1 overexpression: a putative diagnostic biomarker in breast cancer patients," *Gene Reports,* vol. 21, pp. 100791, 2020.

[47] F. Tan, J. Chen, Z. Du, F. Zhao, Y. Liu, Q. Zhang, and C. Yuan, "MIR17HG: A Cancerogenic Long-Noncoding RNA in Different Cancers," *Current Pharmaceutical Design,* vol. 28, no. 15, pp. 1272-1281, 2022.

[48] Y. Liu, F. Zhao, F. Tan, L. Tang, Z. Du, J. Mou, G. Zhou, and C. Yuan, "HNF1A-AS1: A tumor-associated long non-coding RNA," *Current Pharmaceutical Design,* vol. 28, no. 21, pp. 1720-1729, 2022.

[49] R. Abdollahzadeh, A. Azarnezhad, S. Paknahad, Y. Mansoori, M. Pirhoushiaran, K. Kanaani, N. Bafandeh, D. Jafari, and J. Tavakkoly-Bazzaz, "A Proposed TUSC7/miR-211/Nurr1 ceRNET Might Potentially be Disturbed by a cer-SNP rs2615499 in Breast Cancer," *Biochemical Genetics,* vol. 60, no. 6, pp. 2200-2225, Dec, 2022.

[50] S. Hao, W. Han, Y. Ji, H. Sun, H. Shi, J. Ma, J. Yip, and Y. Ding, "BANCR positively regulates the HIF-1α/VEGF-C/VEGFR-3 pathway in a hypoxic microenvironment to promote lymphangiogenesis in pancreatic cancer cells," *Oncology Letters,* vol. 24, no. 6, pp. 1-8, 2022.

[51] X. Wu, T. Xia, M. Cao, P. Zhang, G. Shi, L. Chen, J. Zhang, J. Yin, P. Wu, and B. Cai, "LncRNA BANCR promotes pancreatic cancer tumorigenesis via modulating MiR-195-5p/Wnt/β-catenin signaling pathway," *Technology in Cancer Research & Treatment,* vol. 18, pp. 1533033819887962, 2019.

[52] X. Han, Y. Wang, R. Zhao, G. Zhang, C. Qin, L. Fu, H. Jin, X. Jiang, K. Yang, and H. Cai, "Clinicopathological Significance and Prognostic Values of Long Noncoding RNA BCYRN1 in Cancer

Patients: A Meta-Analysis and Bioinformatics Analysis," *Journal of Oncology,* vol. 2022, 2022.

[53] Y. Li, X. Yang, X. Kang, and S. Liu, "The regulatory roles of long noncoding RNAs in the biological behavior of pancreatic cancer," *Saudi Journal of Gastroenterology,* vol. 25, no. 3, pp. 145-151, May-Jun, 2019.

[54] S. Song, W. Yu, S. Lin, M. Zhang, T. Wang, S. Guo, and H. Wang, "LncRNA ADPGK-AS1 promotes pancreatic cancer progression through activating ZEB1-mediated epithelial-mesenchymal transition," *Cancer Biology & Therapy,* vol. 19, no. 7, pp. 573-583, 2018, 2018.

[55] T. Liu, X. Zhang, Y.-m. Yang, L.-t. Du, and C.-x. Wang, "Increased expression of the long noncoding RNA CRNDE-h indicates a poor prognosis in colorectal cancer, and is positively correlated with IRX5 mRNA expression," *Oncotargets and Therapy,* vol. 9, pp. 1437-1448, 2016, 2016.

[56] X. Ge, Y. Chen, X. Liao, D. Liu, F. Li, H. Ruan, and W. Jia, "Overexpression of long noncoding RNA PCAT-1 is a novel biomarker of poor prognosis in patients with colorectal cancer," *Medical Oncology,* vol. 30, no. 2, pp. 588, Jun, 2013.

[57] B. Liu, Y. Liu, M. Zhou, S. Yao, Z. Bian, D. Liu, B. Fei, Y. Yin, and Z. Huang, "Comprehensive ceRNA network analysis and experimental studies identify an IGF2-AS/miR-150/IGF2 regulatory axis in colorectal cancer," *Pathology-Research and Practice,* vol. 216, no. 10, pp. 153104, 2020.

[58] M.-L. Ma, H.-Y. Zhang, S.-Y. Zhang, and X.-L. Yi, "LncRNA CDKN2B-AS1 sponges miR-28-5p to regulate proliferation and inhibit apoptosis in colorectal cancer," *Oncology Reports,* vol. 46, no. 4, pp. 1-11, 2021.

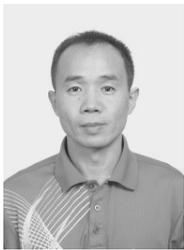

**Jin-Xing Liu** (Member, IEEE) received his B.S. degree in Electronic Information and Electrical Engineering from Shandong University, Jinan, China, in 1997; his M.S. degree in Control Theory and Control Engineering from Qufu Normal University, Jining, China, in 2003; and his Ph.D. degree in Computer Simulation and Control from the South China University of Technology, Guangzhou, China, in 2008. He is a Professor with the School of Health and Life Sciences, University of Health and Rehabilitation Sciences, Qingdao, Shandong. His research interests include pattern recognition, machine learning, and bioinformatics.

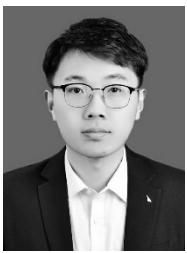

Wen-Yu Xi is a graduate student and the B.S. degree candidate in school of computer Science from QuFu Normal University, China. His research interests include feature selection, principal component analysis, pattern recognition and bioinformatics.

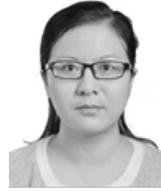

Ling-Yun Dai received the B.S. degree in physics department from Shandong Teacher's University, Jinan, in 2001 and the M.S. degree in School of Information Science and Engineering from Shandong University, Jinan, in 2004. Now she is an associate professor at School of Computer Science, Qufu Normal University, Rizhao. Her research interests include pattern recognition, machine learning, and bioinformatics.

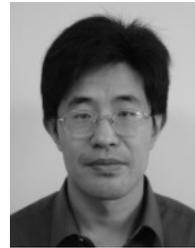

**Chun-Hou Zheng** (Member, IEEE) received his B.S. degree in physics education and the M.S. degree in Control Theory and Control Engineering from Qufu Normal University, China, in 1995 and 2001, respectively; and his Ph.D. degree in Pattern Recognition and Intelligent Systems from the University of Science and Technology of China in 2006. He is currently with the College of Computer Science and Engineering, Anhui University, Hefei, Anhui, China; and with the School of Computer Science, Qufu Normal University, Rizhao, China. His research interests include pattern recognition and bioinformatics.

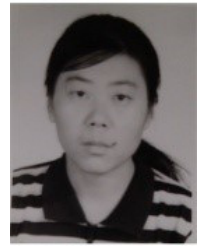

**Ying-Lian Gao** received her B.S. and M.S. degrees from Qufu Normal University, Rizhao, China, in 1997 and 2000, respectively. She is currently with Qufu Normal University Library, Qufu Normal University. Her current research interests include data mining and pattern recognition.